\documentclass[a4paper]{llncs}
\usepackage{amssymb}
\usepackage[export]{adjustbox}
\usepackage{graphicx}
\usepackage{amsmath}
\usepackage[dvipsnames]{xcolor}
\usepackage{multicol}
\usepackage{booktabs}
\usepackage{wrapfig}

\usepackage{algorithm} 
\usepackage[noend]{algpseudocode}

\usepackage{url}
\urldef{\mails}\path|{e.egorov, r.kostoev, e.burnaev}@skoltech.ru, k.necludov@gmail.com|

\usepackage{lineno}

\usepackage{xspace}
\makeatletter
\DeclareRobustCommand\onedot{\futurelet\@let@token\@onedot}
\def\@onedot{\ifx\@let@token.\else.\null\fi\xspace}

\makeatother

\begin{document}

\title{MaxEntropy Pursuit Variational Inference}
\author{Evgenii Egorov\inst{1} \and Kirill Neklydov\inst{2,3} \and Ruslan Kostoev\inst{1} \and Evgeny Burnaev\inst{1}}
\institute{Skolkovo Institute of Science and Technology, Moscow, Russia \and
National Research University Higher School of Economics, Moscow, Russia \and
Samsung AI Center, Moscow, Russia
\mails}

\maketitle

\begin{abstract}
One of the core problems in variational inference is a choice of approximate posterior distribution. It is crucial to trade-off between efficient inference with simple families as mean-field models and accuracy of inference. We propose a variant of a greedy approximation of the posterior distribution with tractable base learners. Using Max-Entropy approach, we obtain a well-defined optimization problem. We demonstrate the ability of the method to capture complex multimodal posterior via continual learning setting for neural networks.

\end{abstract}

\section{Introduction}
\label{sec1}

The posterior distribution evaluation is the primary challenge in Bayesian model construction. Calculating the exact posterior distribution is intractable, and methods like MCMC while being flexible can also be unacceptably expensive. In turn, the variational inference is a method to approximate complicated probability distributions with the simpler ones. Now variational inference is used in semi-supervised classification, drives the most realistic generative models of images, and is a useful tool for analysis of any dynamical system. Inference requires that intractable posterior distributions be approximated by a class of known probability distributions, over which we search for the best representative of the chosen family.

We study the problem of the posterior approximation by a sequentially fitting composition of simple distributions given that one can turn the considered problem to the tractable optimization problem. The structure of the resulting model makes the work with the posterior approximation efficient.

The rest of the paper is organized in the following way. In Section \ref{sec2}, we review the variation inference framework. In Section \ref{sec3}, we derive the stochastic optimization algorithm for sequential approximation of posterior distribution, named MaxEntropy Pursuit Variational Inference. In Section \ref{sec4}, we apply the proposed approach to incremental learning of neural networks. In Section \ref{sec5}, we discuss the obtained results and future work.

\textbf{Notations}. We denote: the differential entropy of distribution $h$ by 
$\mathcal{H}[h] := - \int h \log h d\theta$; the inner product between two Lebesgue integrable functions by $\langle f_1, f_2 \rangle := \int f_1 f_2 d\theta$; the full likelihood of the probabilistic model over the dataset $X$ by $L(\theta) := p(X|\theta)p(\theta)$; the posterior distribution by $p(\theta|X)\propto L(\theta)$.

\section{Variational Inference}
\label{sec2}

We consider the posterior distribution of latent variables $\theta$ given observations $X$:

\begin{equation*}\label{eq:posteriordistribution}
    p(\theta | X) = \dfrac{p(X, \theta)}{\int p(X, \theta) d\theta}.
\end{equation*}
The integral in the denominator is high dimensional, so the normalization is intractable.

The idea of the Variational Inference is to introduce some variational distribution $q_{\lambda}(\theta)$, and instead of computing the normalization constant we approximate the posterior with the simpler distribution $q$, parametrized by the variational parameter $\lambda$ to get the best matching with $p$.

One of the most common approaches to evaluate proximity between $p$ and $q$ is to use KL-divergence (also known as relative entropy or information gain):
\begin{equation*}\label{eq:kldivergence}
    D_{KL}(q(\theta) || p(\theta)) = -\int q(\theta) \log\dfrac{p(\theta)}{q(\theta)}d\theta.
\end{equation*}
KL-divergence is asymmetric ($D_{KL}(q || p) \neq D_{KL}(p || q)$), non-negative and equals to zero iff $q(\theta) = p(\theta)$.

KL-divergence asymmetry provides two different approximation methods: variational inference and expectation propagation (not reviewed in this paper).

Reducing KL-divergence to zero leads to exact matching of distributions, but usually, the variational family $q\in Q$ is not flexible enough for this.

We can formulate minimization of KL-divergence in another way:
\begin{equation*}\label{eq:elbo}
    \begin{aligned}
    & \log{p(X)} = \log{\int p(X, \theta) d\theta} = \log{\int \dfrac{p(X,\theta)q_{\lambda}(\theta)}{q_{\lambda}(\theta)}d\theta} =\\
    & = \log{\mathbb{E}_{q_{\lambda}(\theta)}\left[\dfrac{p(X, \theta)}{q_{\lambda}(\theta)}\right]} \geq \mathbb{E}_{q_{\lambda}(\theta)}\left[\log{\dfrac{p(X, \theta)}{q_{\lambda}(\theta)}}\right] = \mathcal{F}[q] =: ELBO.
    \end{aligned}
\end{equation*}
ELBO (Evidence Lower Bound) with the KL divergence between the
variational distribution and the posterior form the true log marginal probability of the data:
\begin{equation*}\label{eq:truelogprobability}
    \log p(X) = \mathcal{F}(\lambda) + D_{KL}(q_{\lambda}(\theta) || p(X, \theta)),
\end{equation*}
so the minimization of KL-divergence is equivalent to the maximization of ELBO.

However, optimizing over a parametric variational family of distributions and getting the optimal solution $q^{*} = \arg\max\limits_{\mathcal{Q}_{\lambda}}\mathcal{F}[q]$ still leads to the approximation gap \cite{cremer2018inference}, equal to $\log p(X) - \mathcal{F}[q^{*}]$. Many papers showed that the choice of the variational family $\mathcal{Q}_{\lambda}$ is important for quality of the variational approximation \cite{burda2015importance,tran2015variational,salimans2015markov}.

There are a number of approaches for reducing the approximation gap. Some of them propose to increase the flexibility of the approximation family, e.g. normalizing flows \cite{rezende2015variational} or hierarchical variational models \cite{ranganath2016hierarchical}. The other research direction explores the idea of incrementally expanding variational family by the additive mixture of tractable base learners \cite{guo2016boosting,miller2017variational}. In \cite{locatello2018boosting} they investigate the theoretical justification of such approach from an optimization perspective. In general, the both approaches are able to capture the multimodality and nonstandard posterior shapes. However, it seems that the incremental learning of the posterior approximation is more promising from the applied point of view, as the additive mixture composes the approximation using simple and easy-to-evaluate building blocks.

Here we address several problems with this approach. Firstly, starting from the Maximum Entropy principle \cite{caticha2004relative}, we obtain a natural regularized optimization problem, instead of the ad-hoc regularization, proposed in other works. This leads to  interesting connections with other fields and allows to use stochastic optimization approaches in contrast to the original boosting approach \cite{guo2016boosting}. We show the ability of the proposed approach to approximate complex posteriors by using Bayesian Neural Networks, which is a data-intensive and challenging task \cite{welling2011bayesian}.
\section{Max Entropy Pursuit Variational Inference}
\label{sec3}

In this section we derive algorithm in which problem of the posterior distribution is solved by additive mixture. Each component is obtained sequentially. Each step consists of the two optimization problems: for new component $h$ and for the corresponding mixture weight $\alpha$. 

\subsection{Optimization over new component $h$}

Consider that we given some approximation of the posterior distribution $q_t$. Our goal is to improve accuracy of the approximation in terms of the KL-divergence $D_{KL}[q_t(\theta)||p(\theta|X)]$ by using the additive mixture: 
\[q_{t+1} = (1-\alpha)q_t + \alpha h,~\alpha\in(0;1),~h\in Q.\]

Hence, using Maximum Entropy Approach \cite{caticha2004relative} we can state the following optimization problem:

\begin{equation}\label{eq:initproblem}
    \begin{aligned}
    & \max\limits_{h\in\mathcal{Q}}\mathcal{H}[h], s.t. \\
    & \mathcal{F}[q_{t+1}] - \mathcal{F}[q_{t}] > 0.
    \end{aligned}
\end{equation}

As the optimization problem in Eq. \eqref{eq:initproblem} is highly non-linear, we propose to follow the framework based on the Frank-Wolfe algorithm \cite{wang2015functional,locatello2018boosting} and consider the constraint as a functional perturbation.

Expanding the $\mathcal{F}[q_{t+1}]$ term, we get

\begin{equation*}\label{eq:mixexpand}
    \begin{aligned}
    & \mathcal{F}[q_{t+1}] = 
    \int [q_t + \alpha (h-q_t)]\left(\log \frac{L(\theta)}{q_t} - \log \left(1+\alpha\frac{h-q_t}{q_t}\right)\right)d\theta = \\
    & = \underbrace{\int q_t\log \frac{L(\theta)}{q_t} d\theta}_{\text{$\mathcal{F}[q_t]$}} + \alpha\int (h-q_t)\left(\log \frac{L(\theta)}{q_t} - \log \left[1+\alpha\frac{h-q_t}{q_t}\right]\right)d\theta \\
    &- \int q_t \log\left(1+\alpha\frac{h-q_t}{q_t}\right) d\theta.
    \end{aligned}
\end{equation*}

Using Taylor expansion, we obtain the constraint in the following form:

\begin{equation*}\label{eq:taylorexp}
    \begin{aligned}
    & \mathcal{F}[q_{t+1}] - \mathcal{F}[q_{t}] = \alpha \left\langle h-q_t, \log \frac{L(\theta)}{q_t} \right\rangle - \alpha ^2 \int \dfrac{(h-q_t)^2}{q_t}d\theta + o\left(\alpha \left\|\tfrac{h-q_t}{q_t}\right\|_2\right).
    \end{aligned}
\end{equation*}

Considering the first order terms, we get the following optimization problem:
\begin{equation}\label{eq:focproblem}
    \begin{aligned}
    & \max\limits_{h\in Q}\mathcal{H}[h] + \lambda \left\langle h, \log\frac{L(\theta)}{q_t} \right\rangle.
    \end{aligned}
\end{equation}

We can perform scalable optimization by the doubly stochastic gradient descent \cite{kingma2013auto,titsias2014doubly}.  
The $\lambda > 0$ is the corresponding Lagrange multiplier of the constraint. Exact solution of the dual problem for the optimal $\lambda$ is intractable. Below we provide some analysis of how the solution depends on $\lambda$. It allows us to propose practically useful heuristic to select a value of $\lambda$.

Note, that retaining only the first order terms corresponds to the ``functional gradient'' of the KL-divergence \cite{guo2016boosting}. However, MaxEntropy approach allows obtaining the natural regularization term. Further, we show that it is critical to obtain a data scalable algorithm and interpret the parameter $\lambda$. Also, in Section \ref{sec4} we  discuss whether the first order terms expansion is enough for high dimensional problems.

\subsection{Analysis of optimization problem for $h$}

To provide the heuristic rule of choosing the $\lambda$, we optimize in Eq. \eqref{eq:focproblem} not over some parametric family $Q$ of base learners $h$, but over all probability densities. As the objective is concave over $h$, we can derive the global optimal of the maximization problem from the first-order conditions:

\begin{equation*}
    \begin{aligned}
    & \dfrac{\delta}{\delta h}\left[\mathcal{H}[h] + \lambda \left\langle h, \log\frac{L(\theta)}{q_t} \right\rangle \right] + \gamma \left(\int h d\theta - 1\right) = 0, \\
    & h^{*} = \left[\dfrac{L(\theta)}{q_t}\right]^{\lambda}\exp(\gamma-1).
    \end{aligned}
\end{equation*}

Hence, the optimal new component has the following form:

\begin{equation}\label{eq:exactsol}
    \begin{aligned}
    & h^{*} = \dfrac{1}{Z(\lambda)}\left[\dfrac{L(\theta)}{q_t}\right]^{\lambda}.
    \end{aligned}
\end{equation}

The solution $h^{*}$ is intractable, as finding the normalization constant $Z = \int \left[\tfrac{L(\theta)}{q_t}\right]^{\lambda} d\theta$ has the same complexity as solving the original problem. Still, as the global optimum is known, instead of the optimization problem in Eq. \eqref{eq:focproblem} we can consider another optimization problem:

\begin{equation}\label{eq:klinsteadfoc}
    \begin{aligned}
    & \min\limits_{h\in Q}D_{KL}\left(h\Big|\Big|\dfrac{1}{Z(\lambda)}\left[\dfrac{L(\theta)}{q_t}\right]^{\lambda}\right).
    \end{aligned}
\end{equation}

The problem \eqref{eq:klinsteadfoc} is a well-known optimization problem for which there are a lot of black-box variational inference (BBVI) solvers, see e.g. \cite{duvenaud2015black,ranganath2014black}. Hence, any practitioner can benefit from our approach without additional significant costs of implementing or reformulating the initial  statistical problem.
Moreover, we could provide intuition for selecting $\lambda$ by establishing a connection with Renyi divergence \cite{li2016renyi} thanks to the analyses of  the form of \eqref{eq:exactsol}. Namely, we consider a parametric mapping in the probability density space:

\begin{equation}\label{eq:temperaturemapping}
    \begin{aligned}
    & T_{\lambda}: p \to \dfrac{p^{\lambda}(\theta)}{\int p^{\lambda}(\theta)d\theta},~\lambda > 0.
    \end{aligned}
\end{equation}

Consider a pair of a uniform distribution $U$ and $p: \mathcal{H}[p]>\mathcal{H}[U]$. We can easily prove that 
\begin{equation}\label{eq:temperature}
    \begin{aligned}
    & D_{KL}(U||p) > D_{KL}(U||T_{\lambda} p),\text{ for } \lambda > 1, \\
    & D_{KL}(U||p) < D_{KL}(U||T_{\lambda} p),\text{ for } \lambda < 1.
    \end{aligned}
\end{equation}

Hence, we can state that for $\lambda > 1$ we obtain a mode-seeking solution and for $\lambda < 1$ we get a mass covering solution. Interestingly, in case of the Renyi divergence optimization in \cite{minka2005divergence} they describe the same behavior for different values of $\alpha$.
Hence, we can refer to $\lambda$ as the temperature and select some annealing schedule for each step of the optimization process to tune $\lambda$.

Let us consider the corner case, i.e. $\lambda = 1$. Then we can rewrite the objective in  \eqref{eq:focproblem}:

\begin{equation}\label{eq:cornercase}
    \begin{aligned}
    & \arg\max\limits_{h\in Q}\mathcal{H}[h] + \left\langle h, \log\frac{L(\theta)}{q_t}\right\rangle = \arg\max\limits_{h\in Q}
    \underbrace{\int h \log \frac{L(\theta)}{h}d\theta}_{\text{term }(1)} \underbrace{- \int h \log q_t d\theta}_{\text{term }(2)}.
    \end{aligned}
\end{equation}

Hence, the term $1$ in \eqref{eq:cornercase} corresponds to the standard optimization objective in case of variational inference \cite{hoffman2013stochastic}. At the same time  the term $2$ in \eqref{eq:cornercase} plays a role of a penalty for the similarity with the current solution $q_t$.

\subsection{Optimization over mixture weight  $\alpha$ corresponding to $h$}

After we obtain the new mixture component $h$ for the current variational approximation $q_t$, we should select the mixture weight $\alpha$ to obtain a new variational approximation as a convex combination: \[q_{t+1}(\theta) = (1-\alpha)q_{t}(\theta)+\alpha h(\theta).\] Hence, let us  state the optimization problem over $\alpha\in(0;1)$:

\begin{equation}\label{eq:alphainitproblem} 
    \begin{aligned}
    \min\limits_{\alpha\in (0;1)}D_{KL}((1-\alpha)q_{t}(\theta)+\alpha h(\theta)||p(\theta|X)).
    \end{aligned}
\end{equation}

Using Taylor expansion we can get the approximation for any $f$-divergence
\cite{wang2018variational} by the Pearson Chi-squared divergence:

\begin{equation*}\label{eq:approxfdiv}
    \begin{aligned}
    & D_{f}(q||p)\approx f''(1)\chi^2 (q||p).
    \end{aligned}
\end{equation*}

Hence, we can re-formulate the approximation problem:

\begin{equation}\label{eq:alphaapproxprob}
    \begin{aligned}
    & \min\limits_{\alpha\in (0;1)}\int \frac{1}{p(\theta|X)}[q_t+\alpha(h-q_t)]^2d\theta.
    \end{aligned}
\end{equation}

Consider the gradient and the hessian of the objective in \eqref{eq:alphaapproxprob} w.r.t. $\alpha$:

\begin{equation*}\label{eq:chigradhess}
    \begin{aligned}
    & \nabla_{\alpha}\int \frac{1}{p}[q_t+\alpha(h-q_t)]^2d\theta = 
    2\int \frac{1}{p(\theta|X)}[q_t+\alpha(h-q_t)](h-q_t)d\theta, \\
    & \nabla^2_{\alpha}\int \frac{1}{p(\theta|X)}[q_t+\alpha(h-q_t)]^2d\theta = 2\int\frac{(h-q_t)^2}{p(\theta|X)} > 0.
    \end{aligned}
\end{equation*}.

As the objective \eqref{eq:alphaapproxprob} is convex, we can obtain the solution of the optimization problem \eqref{eq:alphaapproxprob} from the first order condition:

\begin{equation}\label{xxxx}
    \begin{aligned}
    & \alpha^{*} = - \dfrac{\int \frac{1}{p(\theta|X)}q_t(h-q_t)d\theta }{\int \frac{1}{p(\theta|X)}(h-q_t)^2d\theta} = - \dfrac{\int \frac{1}{L(\theta)}q_t(h-q_t)d\theta }{\int \frac{1}{L(\theta)}(h-q_t)^2d\theta}.
    \end{aligned}
\end{equation}

In practice such estimator has high variance. Estimation for each sample requires the forward pass through the whole dataset, hence the variance can not be reduced by averaging efficiently. 
Therefore we propose to use the exact solution \eqref{xxxx} in case of middle-size datasets and use the stochastic gradient approach with a projection for the objective from Eq. \eqref{eq:alphainitproblem} in case of large-scale datasets.
\section{Neural Network Incremental Learning via Bayesian Inference}
\label{sec4}

Deep neural networks provide the state-of-art solution for the image classification problems. However, as a network is trained to do a specific classification task, it is problematic to incrementally learn any new task. This situation was described as the catastrophically forgetting behaviour of neural networks. However, intuitively we expect the other situation: performance should similar to that when training over the whole dataset in the offline mode \cite{kemker2018measuring}. In this section, we show how our approach helps to overcome this limitation.

\begin{figure}[!h]
    \centering
    \includegraphics[scale=0.6]{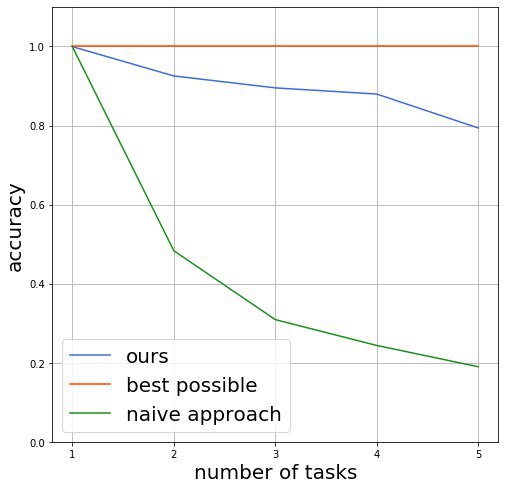}
    \caption{Mean test accuracy for a sequence of models, trained on a sequence of tasks (subsets of the training set). Each next task is equal to a previous task plus some new subset of the initial training data.}
    \label{fig:tasks}
\end{figure}

\textbf{Experimental Setup}. We perform the incremental class learning experiment using the MNIST dataset with the LeNet-5 Convolutional Neural Networ (CNN) \cite{lecun1998gradient}. The dataset contains grey scale images belonging to 10 classes. We split the dataset in 5 tasks, the first task containing digits '0' and '1', the second task containing digits '2' and '3', and so on. For each task, we perform 10 epoch of training. We compare our incremental posterior approximation of the neural network parameters with a baseline naive continual neural network learning. The size of the test dataset is $10^4$ samples, the total train size for all tasks is $5\times 10^4$. As the prior distribution on the neural network parameters we use the fully factorized standard normal distribution. The predictive distribution of the model is approximated by an ensemble of the weights sampled from the variational approximation.

\textbf{Results}. As result, we find our incremental posterior distribution approximation to maintain higher test accuracy through the whole sequence of tasks, almost matching the performance of a network trained simultaneously on all observed data. Fig. \ref{fig:tasks} shows the test accuracy as new tasks are observed. We conclude from our results that the incremental posterior approximation leads to a drastic increase in performance for incremental learning tasks. 

%

\section{Conclusion}
\label{sec5}

In this work, we developed an efficient approach for learning complex multimodal posteriors by constructing an additive mixture of simple densities. Following the MaxEntropy approach, we state  well defined and tractable optimization problem. Additive mixture allows us to control the complexity of the posterior by simply increasing or decreasing the number of components. 

An important avenue of future research is to develop approaches for modeling covariance structure that accurately account for different characteristics of the posterior and that still allow for efficient computations in case of deep neural networks.

Also, we plan to consider various applications of the proposed approximation scheme including uncertainty quantification \cite{ADoESobol2015,ADoESobol2016,ADoESobol2017} and Bayesian parameter estimation for Gaussian Processes regression \cite{GPBvM2013a,GPBvM2013b,GPBvM2014c}.

\section*{Acknowledgements}

The work was supported by the Russian Science Foundation under Grant 19-41-04109.

\clearpage
\newpage

{\small
\bibliographystyle{splncs04}
\bibliography{references}
}

\end{document}